\documentclass{article}

    \usepackage[final]{neurips_2020}

\usepackage[utf8]{inputenc} 
\usepackage[T1]{fontenc}    
\usepackage{hyperref}       
\usepackage{url}            
\usepackage{booktabs}       
\usepackage{amsfonts}       
\usepackage{nicefrac}       
\usepackage{microtype}      
\usepackage{xcolor}
\usepackage{multirow}
\usepackage{multicol}
\usepackage{arydshln}
\usepackage{amsmath}
\usepackage{graphicx}
\usepackage{caption}
\usepackage{subcaption}
\usepackage{bm}
\usepackage{bbm}
\usepackage{lipsum}
\usepackage{wrapfig}
\usepackage[ruled, linesnumbered]{algorithm2e}
\usepackage[titletoc]{appendix}
\definecolor{mydarkblue}{rgb}{0,0.08,0.45}
\hypersetup{
    colorlinks   = true,
    citecolor    = mydarkblue
}
\newcommand{\bftab}{\fontseries{b}\selectfont}
\DeclareMathOperator*{\argmax}{arg\,max}

\title{Model-based Reinforcement Learning for Semi-Markov Decision Processes with Neural ODEs}

\author{%
  Jianzhun Du,\quad Joseph Futoma,\quad Finale Doshi-Velez \\
  Harvard University\\
  Cambridge, MA 02138 \\
  \texttt{jzdu@g.harvard.edu, \{jfutoma, finale\}@seas.harvard.edu} \\
}

\begin{document}

\maketitle

\vspace{-0.35cm}
\begin{abstract}
We present two elegant solutions for modeling continuous-time dynamics, in a novel model-based reinforcement learning (RL) framework for semi-Markov decision processes (SMDPs) using neural ordinary differential equations (ODEs). Our models accurately characterize continuous-time dynamics and enable us to develop high-performing policies using a small amount of data. We also develop a model-based approach for optimizing time schedules to reduce interaction rates with the environment while maintaining the near-optimal performance, which is not possible for model-free methods. We experimentally demonstrate the efficacy of our methods across various continuous-time domains.
\end{abstract}

\section{Introduction}\label{sec:intro}
Algorithms for deep reinforcement learning (RL) have led to major advances in many applications ranging from robots \citep{gu2017deep} to Atari games \citep{mnih2015human}.  Most of these algorithms formulate the problem in discrete time and assume observations are available at every step.  However, many real-world sequential decision-making problems operate in continuous time. For instance, control problems such as robotic manipulation are generally governed by systems of differential equations.  In healthcare, patient observations often consist of irregularly sampled time series: more measurements are taken when patients are sicker or more concerns are suspected, and clinical variables are usually observed on different time-scales.

Unfortunately, the problem of learning and acting in continuous-time environments has largely been passed over by the recent advances of deep RL.  Previous methods using semi-Markov decision processes (SMDPs) \citep{smdp}---including $Q$-learning \citep{bradtke1995reinforcement}, advantage updating \citep{baird1994reinforcement}, policy gradient \citep{munos2006policy}, actor-critic \citep{doya2000reinforcement}---extend the standard RL framework to continuous time, but all use relatively simple linear function approximators. Furthermore, as \emph{model-free} methods, they often require large amounts of training data. Thus, rather than attempt to handle continuous time directly, practitioners often resort to discretizing time into evenly spaced intervals and apply standard RL algorithms. However, this heuristic approach loses information about the dynamics if the discretization is too coarse, and results in overly-long time horizons if the discretization is too fine.

In this paper, we take a \emph{model-based} approach to continuous-time RL, modeling the dynamics via neural ordinary differential equations (ODEs) \citep{chen2018neural}.  Not only is this more sample-efficient than model-free approaches, but it allows us to efficiently adapt policies learned using one schedule of interactions with the environment for another. Our approach also allows for optimizing the measurement schedules to minimize interaction with the environment while maintaining the near-optimal performance that would be achieved by constant intervention.

Specifically, to build flexible models for continuous-time, model-based RL, we first introduce ways to incorporate \emph{action} and \emph{time} into the neural ODE work of \citep{chen2018neural, rubanova2019latent}. We present two solutions, ODE-RNN (based on a recurrent architecture) and Latent-ODE (based on an encoder-decoder architecture), both of which are significantly more robust than current approaches for continuous-time dynamics. Because these models include a hidden state, they can handle partially observable environments as well as fully-observed environments. Next, we develop a unified framework that can be used to learn both the \emph{state transition} and the \emph{interval timing} for the associated SMDP. Not only does our model-based approach outperform baselines in several standard tasks, we demonstrate the above capabilities which are not possible with current model-free methods.

\section{Related work}

There has been a large body of work on continuous-time reinforcement learning, many based on the RL framework of SMDPs \citep{bradtke1995reinforcement, parr1998hierarchical}. Methods with linear function approximators include $Q$-learning \citep{bradtke1995reinforcement}, advantage updating \citep{baird1994reinforcement}, policy gradient \citep{munos2006policy}, and actor-critic \citep{doya2000reinforcement}.  Classical control techniques such as the linear–quadratic regulator (LQR) \citep{kwakernaak1972linear} also operate in continuous time using a linear model class that is likely too simplistic and restrictive for many real-world scenarios.  We use a more flexible model class for learning continuous-time dynamics models to tackle a wider range of settings.

Other works have considered SMDPs in the context of varying time discretizations.  Options and hierarchical RL \citep{sutton1999between, barto2003recent} contain temporally extended actions and meta-actions.  More recently, \citet{sharma2017learning} connected action repetition of deep RL agents with SMDPs, and \citet{tallec2019making} identified the robustness of $Q$-learning on different time discretizations; however, the transition times were still evenly-spaced.  In contrast, our work focuses on truly continuous-time environments with irregular, discrete intervention points.

More generally, discrete-time, model-based RL has offered a sample-efficient approach \citep{kaelbling1996reinforcement} for real-world sequential decision-making problems. Recently, RNN variants have become popular black-box methods for summarizing long-term dependencies needed for prediction. RNN-based agents have been used to play video games \citep{oh2015action, chiappa2017recurrent}; \citet{ha2018world} trained agents in a ``dreamland'' built using RNNs; \citet{igl2018deep} utilized RNNs to characterize belief states in situations with partially observable dynamics; and \citet{neitz2018adaptive} trained a recurrent dynamics model  skipping observations adaptively to avoid poor local optima. To our knowledge, no prior work in model-based RL focuses on modeling continuous-time dynamics and planning with irregular observation and action times.

Neural ODEs \citep{chen2018neural} have been used to tackle irregular time series. \citet{rubanova2019latent, de2019gru} used a neural ODE to update the hidden state of recurrent cells; \citet{chen2018neural} defined latent variables for observations as the solution to an ODE; \citet{kidger2020neuralcde} adjusted trajectories based on subsequent observations with controlled differential equations. To our knowledge, ours is the first to extend the applicability of neural ODEs to RL. 

\section{Background and notation} 

\paragraph{Semi-Markov decision processes.}
A semi-Markov decision process (SMDP) is a tuple $(\mathcal{S}, \mathcal{A}, \mathcal{P}, \mathcal{R}, \mathcal{T}, \gamma)$, where $\mathcal{S}$ is the state space, $\mathcal{A}$ is the action space, $\mathcal{T}$ is the transition time space and $\gamma \in (0,1]$ is the discount factor. We assume the environment has transition dynamics $P(s', \tau|s,a) \in \mathcal{P}$ unknown to the agent, where $\tau$ represents the time between taking action $a$ in observed state $s$ and arriving in the next state $s'$ and can take a new action. Thus, we assume no access to any intermediate observations. In general, we are given the reward function $r = R(s,a,s')$ for the reward after observing $s'$. However, in some cases the reward may also depend on $\tau$, i.e. $r = R(s,a,s',\tau)$, for instance if the cost of a system involves how much time has elapsed since the last intervention. The goal throughout is to learn a policy maximizing long-term expected rewards $\mathbb{E}\left[\sum_{i=1}^L \gamma^{t_i} r_i\right]$ with a finite horizon $T=t_L$, where $t_i = \sum_{j=1}^i \tau_j$. 

While the standard SMDP model above assumes full observability, in our models, we will introduce a latent variable $z \in \mathbb{R}^v$ summarizing the history until right before the most recent state, and learn a transition function $\hat{P}(z',\tau|z,a,s)$, treating $s$ as an \emph{emission} of the latent $z$.  Introducing this latent $z$ will allow us to consider situations in which: (a) the state $s$ can be compressed, and (b) we only receive partial observations and not the complete state with one coherent model. 

\paragraph{Neural ordinary differential equations.} 
Neural ODEs define a latent state $z(t)$ as the solution to an ODE initial value problem using a time-invariant neural network $f_\theta$: 
\begin{gather}
    \frac{d z(t)}{dt} = f_\theta (z(t), t), \quad \text{where} \; z(t_0) = z_0.
\end{gather}
Utilizing an off-the-shelf numerical integrator, we can solve the ODE for $z$ at any desired time. In this work, we consider two different neural ODE models as starting points.  First, a standard RNN can be transformed to an ODE-RNN \citep{rubanova2019latent}:
\begin{equation}
    \begin{gathered}
        \tilde{z}_{i-1} = \text{ODESolve}\left(f_\theta, z_{i-1}, \tau_{i-1} \right), \quad z_i = \text{RNNCell}\left(\tilde{z}_{i-1}, s_{i-1}\right), \quad \hat{s}_i = o(z_i).
    \end{gathered}
    \label{eq:vanilla-ode-rnn}
\end{equation}
where $\hat{s}$ is the predicted state and $o(\cdot)$ is the emission (decoding) function. An alternate approach, based on an encoder-decoder structure \citep{sutskever2014sequence}, is the Latent-ODE \citep{chen2018neural}: 
\begin{equation}
    \begin{gathered}
        z_0 \sim q_\phi(z_0|\{s_0\}_{i=0}^L), \quad \{z_i\}_{i=1}^L = \text{ODESolve}\left(f_\theta, z_0, \{\tau_0\}_{i=0}^{L-1} \right), \quad \hat{s}_i = o(z_i),
    \end{gathered}
    \label{eq:vanilla-latent-ode}
\end{equation}
where $q_\phi$ is a RNN encoder and the latent state $z$ is defined by an ODE. The Latent-ODE is trained as a variational autoencoder (VAE) \citep{kingma2013auto, rezende2014stochastic}. 

The ODE-RNN allows online predictions and is natural for sequential decision-making, though the effect of the ODE network is hard to interpret in the RNN updates. On the other hand, the Latent-ODE explicitly models continuous-time dynamics using an ODE, along with a measure of uncertainty from the posterior over $z$, but the solution to this ODE is determined entirely by the initial latent state. 

\paragraph{Recurrent environment simulator.}
None of the above neural ODE models contain explicit actions. We build on the recurrent simulator of \citet{chiappa2017recurrent}, which used the following structure for incorporating the effects of actions:
\begin{equation}
    \begin{gathered}
        z_i = \text{RNNCell}\left(z_{i-1}, a_{i-1}, \dot{s}_i\right), \quad \hat{s}_i = o(z_i),
    \end{gathered}
    \label{eq:rnn-transition}
\end{equation}
where $\dot{s}$ denotes either the observed state $s$ or the predicted state $\hat{s}$. We can use either $s$ or $\hat{s}$ during training of the recurrent simulator, but only $\hat{s}$ is available for inference. In this work, we generally use the actual observations during training, i.e. $\dot{s}=s$ (this is known as the teacher forcing strategy), but also find that using \emph{scheduled sampling} \citep{bengio2015scheduled} which switches between choosing the previous observation and the prediction improves performance on some tasks.

\section{Approach}
In this section, we first describe how to construct ODE-based dynamics models for model-based RL that account for both actions and time intervals, overcoming shortcomings of existing neural ODEs.  Then, we describe how to use these models for prediction in the original environment, as well as how to transfer to environments with new time schedules and how to optimize environment interaction.

\subsection{Model definition and learning}\label{sec:model}
We decompose the transition dynamics into two parts, one to predict the time $\tau$ until the next action with corresponding observation, and one to predict the next latent state $z'$:
\begin{gather}
    P(z', \tau|z,a,s) = P(z'|z,a,s,\tau) \cdot P(\tau|z,a,s). 
\end{gather}
Using the recurrent environment simulator from Equation \ref{eq:rnn-transition} we can incorporate an action $a$ into an ODE-RNN or a Latent-ODE to approximate $P(z'|z,a,s,\tau)$ (referred to as the \emph{state transition}). We can also learn and generate transition times $\tau$ using another neural network $g_\kappa(z,a,s)$ based on the current latent state, action and observed state, which approximates $P(\tau|z,a,s)$ (referred to as the \emph{interval timing}). We model them separately yet optimize jointly in a multi-task learning fashion. Specifically, we propose an \emph{action-conditioned} and \emph{time-dependent} ODE-RNN and Latent-ODE for approximating the transition dynamics $P(z', \tau|z,a,s)$. 

\paragraph{ODE-RNN.} Combining Equations \ref{eq:vanilla-ode-rnn} and \ref{eq:rnn-transition}, we obtain the following model:
\begin{equation}
   \begin{gathered}
        \tilde{z}_{i-1} = \text{ODESolve}\left(f_\theta, z_{i-1}, \dot{\tau}_{i-1} \right), \\
        z_i = \text{RNNCell}\left(\tilde{z}_{i-1}, a_{i-1}, \dot{s}_{i-1} \right), \\
        \hat{s}_i = o(z_i),\quad
        \hat{\tau}_i = g_\kappa(z_i, a_i, \dot{s}_i),
    \end{gathered} 
    \label{eq:ode-rnn}
\end{equation}
where $\dot{\tau}$ denotes either the observed time interval $\tau$ or the predicted time interval $\hat{\tau}$, similar to $\dot{s}$. 

\begin{figure}[t]
    \centering
    \includegraphics[width=0.95\linewidth]{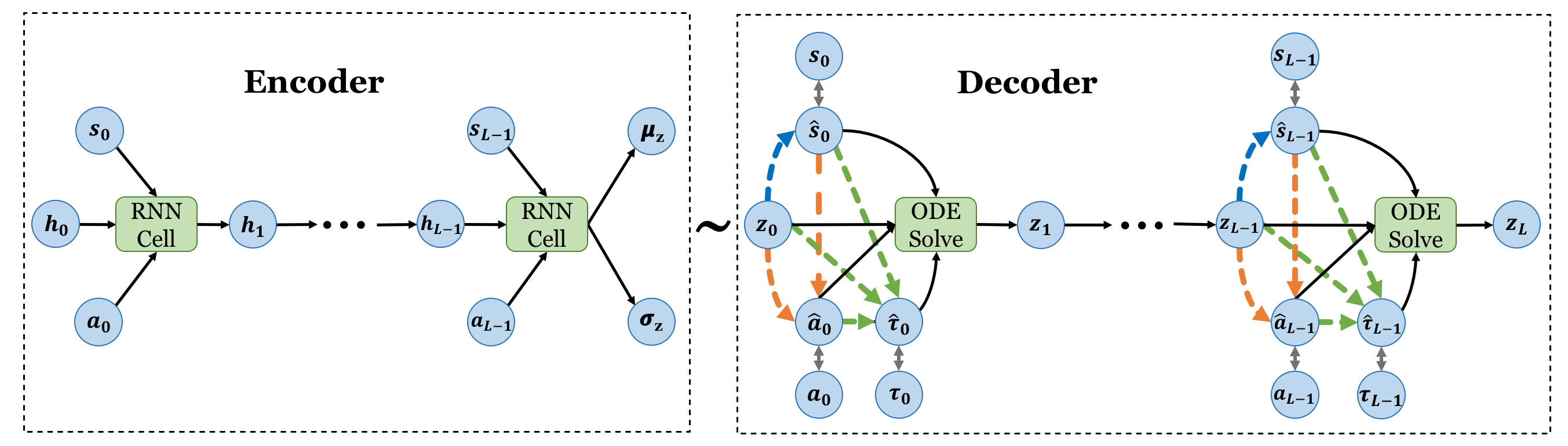}
    \caption{The graphical representation of action-conditioned and time-dependent Latent ODE.
    \textcolor{blue}{Blue} dashed arrow represents the emission function. \textcolor{orange}{Orange} dashed arrow represents the (latent) policy. \textcolor{green}{Green} dash arrow represents the prediction of interval timing. \textcolor{gray}{Gray} double-sided arrow represents the selection between observation and prediction. Note that the encoder is only used in model training.}
    \label{fig:latent-ode}
\end{figure}

\paragraph{Latent-ODE.} 
Given the parameters $\theta$ of the underlying dynamics, the entire latent trajectory of the vanilla Latent-ODE is determined by the initial condition, $z_0$. However, to be useful as a dynamics model for RL the Latent-ODE should allow for actions to modify the latent state. To do this, at every time we adjust the previous latent state to obtain a new latent state $\tilde{z}$, incorporating the new actions and observations. In particular, we transform the vanilla Latent-ODE using Equations \ref{eq:vanilla-latent-ode} and \ref{eq:rnn-transition}:
\begin{equation}
    \begin{gathered}
        z_0 \sim q_\phi(z_0|s_0, a_0, s_1, a_1, \ldots, s_{L-1}, a_{L-1}), \\
        \tilde{z}_{i-1} = \xi_\psi(z_{i-1}, a_{i-1}, \dot{s}_{i-1}), \\
        z_i = \text{ODESolve}\left(f_\theta, \tilde{z}_{i-1}, \dot{\tau}_{i-1} \right), \\
        \hat{s}_i = o(z_i), \quad
        \hat{\tau}_i = g_\kappa(z_i, a_i, \dot{s}_i),
    \end{gathered}
    \label{eq:latent-ode}
\end{equation}
where $\xi_\psi: \mathbb{R}^w \to \mathbb{R}^v$ is a function incorporating the action and observation (or prediction) to the ODE solution. This method of incorporating actions into neural ODEs is similar to \emph{neural controlled differential equations} \citep{kidger2020neuralcde}. 
We find that a linear transformation works well in practice for $\xi_\psi$, i.e.,
\begin{gather}
    \tilde{z}_{i-1} = W [z_{i-1}\; a_{i-1}\; \dot{s}_{i-1}] + b.
\end{gather}
where $[\cdot]$ is the vector concatenation. The graphical model of Latent-ODE is shown in Figure \ref{fig:latent-ode}.

The ODE-RNN and Latent ODE intrinsically differ in the main entity they use for modeling continuous-time dynamics. The ODE-RNN models the transition between latent states using a recurrent unit, whereas the Latent-ODE parameterizes the dynamics with an ODE network directly. 

\paragraph{Training objective.}  
We assume that we have a collection of variable-length trajectories $h = \left(\{s^{(n)}_i\}_{i=0}^{L_n}, \{a^{(n)}_i\}_{i=0}^{L_n-1}, \{\tau^{(n)}_i\}_{i=0}^{L_n-1}\right)_{n=1}^N$. We optimize the overall objective in Equation \ref{eq:loss} using mini-batch stochastic gradient descent:
\begin{gather}
    \mathcal{L}(h) = \mathcal{L}^\mathcal{S}(h) + \lambda\cdot\mathcal{L}^\mathcal{T}(h),
    \label{eq:loss}
\end{gather}
where $\mathcal{L}^\mathcal{S}(h)$ is a loss for prediction of state transitions, $\mathcal{L}^\mathcal{T}(h)$ is a loss for prediction of interval timings, and $\lambda$ is a hyperparameter trades off between the two. Based on the complexity or importance of predicting state transitions and interval timings, we can emphasize either one by adjusting $\lambda$. 

For recurrent models, such as the ODE-RNN, $\mathcal{L}^\mathcal{S}(h)$ is simply the mean squared error (MSE); for encoder-decoder models, such as the Latent ODE, it is the negative evidence lower bound (ELBO):

\begin{equation}
    \begin{gathered}
        \mathcal{L}^\mathcal{S}(h) =
        \begin{cases}
            \frac{1}{N \sum_{n=1}^N L_n}\sum_{n=1}^N \sum_{i=1}^{L_n} \|\hat{s}^{(n)}_i - s^{(n)}_i\|_2^2 \hspace{2.35cm} \text{for recurrent models}, \\
            -\mathbb{E}_{z_0^{(n)} \sim q_\phi\left(z_0^{(n)} | \{s^{(n)}_i, a^{(n)}_i\}_{i=0}^{L_n-1}\right)} \left[\log p(\{s^{(n)}_i\}_{i=1}^{L_n} \big| s_0, \{z^{(n)}_i, a^{(n)}_i, \tau^{(n)}_i\}_{i=0}^{L_n-1})\right] \\
            \hspace{0.5cm}+\mathbb{KL}\left[q_\phi\left(z_0^{(n)} \big| \{s^{(n)}_i, a^{(n)}_i\}_{i=0}^{L_n-1}\right) \| p(z_0^{(n)})\right] \hspace{1cm} \text{for encoder-decoder models}.
        \end{cases}
    \end{gathered} 
    \label{eq:loss-state}
\end{equation}

For the loss $\mathcal{L}^\mathcal{T}(h)$ for interval timing (Equation \ref{eq:loss-time}), we use cross entropy for a small number of discrete $\tau$. Otherwise, MSE can be used for continuous $\tau$.
\begin{equation}
    \begin{gathered}
        \mathcal{L}^\mathcal{T}(h) =
        \begin{cases}
            -\frac{1}{N}\sum_{n=1}^N \sum_{i=0}^{L_n-1} \sum_{m=1}^M y^{(n)}_{i, m} \log p^{(n)}_{i, m} & \text{for classification}, \\
            \frac{1}{N\sum_{n=1}^N L_n}\sum_{n=1}^N \sum_{i=0}^{L_n-1} \|\hat{\tau}^{(n)}_i - \tau^{(n)}_i\|_2^2 & \text{for regression},
        \end{cases}
    \end{gathered} 
    \label{eq:loss-time}
\end{equation}
where $M$ is the number of classes of time interval, $y^{(n)}_{i, m}$ is the binary indicator if the class label $m$ is the correct classification for $\tau^{(n)}_i$, and $p^{(n)}_{i, m}$ is the predicted probability that $\tau^{(n)}_i$ is of class $m$.

\subsection{Planning and learning}\label{sec:planning}
Now that we have models and procedures for learning them, we move on to the question of identifying an optimal policy. With partial observations, the introduced latent state $z$ provides a representation encoding previously seen information \citep{ha2018world} and we construct a \emph{latent policy} $\pi_\omega(a|s, z)$ conditioned on $z$; otherwise the environment is fully observable and we construct a policy $\pi_\omega(a|s)$. In general, we model the action-value function $Q(s, a)$ (or $Q([s\;z], a)$) with continuous-time $Q$-learning \citep{bradtke1995reinforcement, sutton1999between} for SMDPs, which works the unequally-sized time gap $\tau$ into the discount factor $\gamma$:
\begin{gather}
    Q(s,a) \leftarrow Q(s,a) + \alpha \left[r + \gamma^\tau \max_{a'} Q(s',a') - Q(s,a)\right].
    \label{eq:q-learning}
\end{gather}
We construct the policy $\pi_\omega$ with a deep $Q$-network (DQN) \citep{mnih2015human} for discrete actions and deep deterministic policy gradient (DDPG) \citep{lillicrap2015continuous} for continuous actions. We perform efficient planning with dynamics models to learn the policy $\pi_\omega$, which is detailed in Section \ref{sec:exp}. The exact method for planning is orthogonal to the use of our models and framework.

\paragraph{Transferring to environments with different time schedules.}
Our model-based approach allows us to adapt to changes in time interval settings: once we have learned the underlying state transition dynamics $P(z'|z,s,a,\tau)$ from a particular time schedule $p(\tau)$, the model can be used to find a policy for another environment with any different time schedules $p'(\tau)$ (either irregular or regular). The model is generalizable for interval times if and only if it truly learns how the system changes over time in the continuous-time environment.

\paragraph{Interpolating rewards and optimizing interval times.}
In addition to adapting to new time schedules, we can also optimize a state-specific sampling rate to maximize cumulative rewards while minimizing interactions. For example, in Section~\ref{sec:demo}, we will demonstrate how we can reduce the number of times a (simulated) HIV patient must come into the clinic for measurement and treatment adjustment while still staying healthy. However, this approach may not work well in situations where constant adjustments on small time scales can hurt performance (e.g., Atari games use frameskipping to avoid flashy animations). 

When optimizing interval times, we assume that $\tau$ are discrete, $\tau \in \{1,2,\dots,\delta\}$, and we only have access to the immediate reward function $r = R(s,a,s')$.  We can optimize the interval times to decrease the amount of interaction with the environment while achieving near-optimal performance, obtained by maximal interaction ($\tau \equiv 1$). Specifically, \emph{assuming we always take an action $a$} for each of the $\tau$ steps in the interval starting from the state $s_0$, we select the optimal $\tau^\ast$ based on the estimated $\tau$-step ahead value:
\begin{gather}
    \tau^\ast = \argmax_{\tau \in [1, 2, \ldots, \delta]} \sum_{i=1}^{\tau} \gamma^{i-1} \hat{r}_i + \gamma^\tau Q^\pi(\hat{s}_\tau, \hat{a}_\tau), \quad \text{where} \; \hat{r}_i = R(\hat{s}_{i-1},a,\hat{s}_i),
    \label{eq:opt}
\end{gather}
$\hat{s}$ is the simulated state from the model ($\hat{s}_0 = s_0$), $\hat{a}$ is the optimal action at state $\hat{s}$, and $Q^\pi$ is the action-value function from the policy. We interpolate intermediate rewards using the dynamics model---we can simulate by varying $\tau$ what states would be passed through, and what the cumulative reward would be---whereas it is not possible to do this for model-free methods as we have no access to in-between observations. In this way, agents learn to skip situations in which no change of action is needed and minimize environment interventions. The full procedure can be found in Appendix \ref{apx:algo-opt}. Note that Equation \ref{eq:opt} can be easily extended to continuous interval times if they are lower-bounded, i.e., $\tau \in [a, \infty)$ ($a \in \mathbb{R}^+$); we leave this to future work.

\section{Experiments}\label{sec:exp}
We evaluate our ODE-based models across four continuous-time domains. We show our models characterize continuous-time dynamics more accurately and allow us to find a good policy with less data. We also demonstrate capabilities of our model-based methods that are not possible for model-free methods.

\subsection{Experimental setup}

\begin{figure}[t]
    \centering
    \begin{subfigure}[t]{0.31\textwidth}
        \centering
        \includegraphics[width=\textwidth]{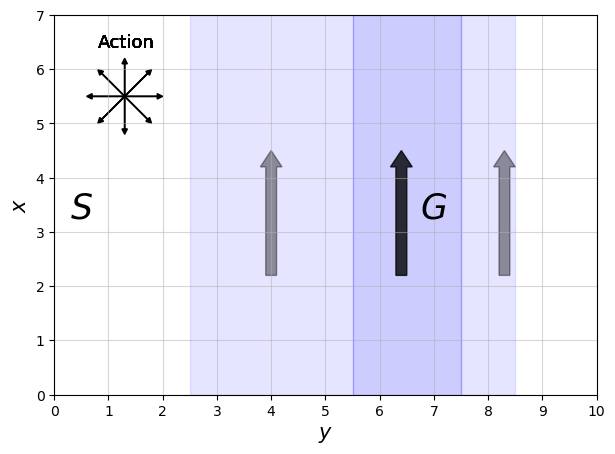}
        \caption{}
        \label{fig:grid}
    \end{subfigure}%
    \hspace{.5cm}
    \begin{subfigure}[t]{0.6\textwidth}
        \centering
        \includegraphics[width=\textwidth]{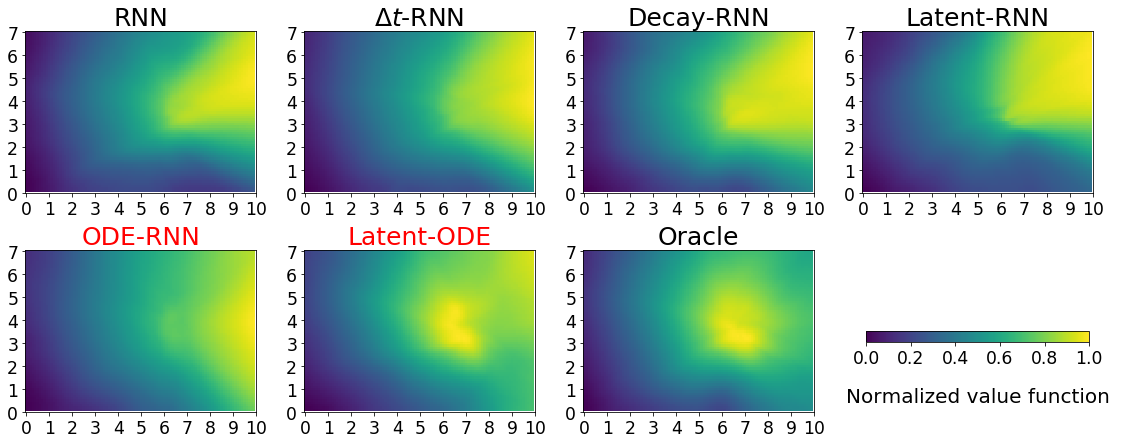}
        \caption{}
        \label{fig:value-func}
    \end{subfigure}
    \caption{(a) 2D continuous windy gridworld, where ``S'' and ``G'' is the start area $(x \in [3, 4], y \in [0, 1])$ and goal area $(x \in [3, 4], y \in [6.5, 7.5])$, and the shaded region and arrows represent the wind moves upward (the darker color indicates the stronger wind); (b) normalized value functions obtained by DQNs for all baselines and our proposed ODE-based models (marked as red) over the gridworld. ``Oracle'' refers to the model-free baseline trained until convergence. The lighter the pixel, the higher the value. The policy developed with the Latent-ODE is ``closet'' to the optimal policy (oracle).}
    \vspace{-0.3cm}
\end{figure}

\paragraph{Domains.}
We provide demonstrations on three simpler domains---windy gridworld \citep{sutton2018reinforcement}, acrobot \citep{sutton1996generalization}, and HIV \citep{adams2004dynamic}---and three Mujoco \citep{todorov2012mujoco} locomotion tasks---Swimmer, Hopper and HalfCheetah---interfaced through OpenAI Gym \citep{1606.01540}. Unless otherwise stated, the state space of all tasks is fully observable, and we are given the immediate reward function $r=R(s,a,s')$. We provide more details of domains in Appendix \ref{apx:env}.
\setlength{\leftmargini}{2.5em}
\begin{itemize}
    \item \textit{Windy gridworld} (Figure \ref{fig:grid}). We consider a continuous state version in which agents pursue actions $a=(\Delta x, \Delta y)$ for $\tau \sim \text{Unif}\{1,7\}$ seconds to reach a goal region despite a crosswind.  
    \item \textit{Acrobot}. Acrobot aims to swing up a pendulum to reach a given height. The dynamics is defined by a set of first-order ODEs and the original version uses $\tau=0.2$ to discretize underlying ODEs. We instead sample the time interval $\tau$ randomly from $\{0.2,0.4,0.6,0.8,1\}$. 
    \item \textit{HIV.} Establishing effective treatment strategies for HIV-infected patients based on markers from blood tests can be cast as an RL problem \citep{ernst2006clinical, parbhoo2017combining, killian2017robust}. The effective period $\tau$ varies from one day to two weeks. Healthy patients with less severe disease status may only need occasional inspection, whereas unhealthy patients require more frequent monitoring and intervention. 
    \item \textit{Mujoco.} We use action repetition \citep{sharma2017learning} to introduce irregular transition times to  Mujoco tasks, however, we assume the intermediate observations are not available so that the dynamics naturally fits into the SMDP definition.
    The same action repeats $\tau = h(\|\theta_v\|_2)$ times where $\theta_v$ is the angle velocity vector of all joints and $h$ is a periodic function. The periodicity is learnable by RNNs \citep{gers2002learning} and ensures consistently irregular measurements in the course of policy learning\footnote{Otherwise $\tau$ remains unchanged after robots learn to run, because $\|\theta_v\|_2$ only fluctuates in a small range.}. 
\end{itemize}
As proofs of concept, we assume the transition times in gridworld and acrobot problems are known, and only focus on learning the state transition probability $P(z'|z,a,s,\tau)$ (set $\lambda=0$ in Equation \ref{eq:loss}); for HIV and Mujoco tasks, we learn both state transitions and interval timings.

\paragraph{Baselines.}
We compare the performance of our proposed ODE-based models with four baselines embedded in our model-based RL framework for SMDPs. With the recurrent architecture, we evaluate 1) vanilla RNN; 2) $\Delta t$-RNN, where the time intervals are concatenated with the original input as an extra feature; 3) Decay-RNN \citep{che2018recurrent}, which adds an exponential decay layer between hidden states: $\tilde{z} = e^{-\max\{0, w\tau + b\}} \odot z$. With the encoder-decoder architecture, we evaluate 4) Latent-RNN, where the decoder is constructed with a RNN and the model is trained variationally. RNNs in all models are implemented by gate recurrent units (GRUs) \citep{cho2014learning}. Moreover, we also run a model-free method (DQN or DDPG) for comparison.

\subsection{Demonstrations on simpler domains}\label{sec:demo}
We learn the \emph{world model} \citep{ha2018world} of simpler environments for planning. We gather data from an initial random policy and learn the dynamics model on this collection of data. Agent policies are \emph{only} trained using fictional samples generated by the model \emph{without considering the planning horizon}. To achieve optimal performance (i.e. the model-free baseline trained until convergence), the model has to capture long-term dependencies so that the created virtual world is accurate enough to mimic the actual environment. Thus, with this planning scheme, we can 
clearly demonstrate a learned model's capacity. The details of the algorithm can be found in Appendix \ref{apx:algo-world-model}. 

\begin{table}[t]
    \scriptsize
    \centering
    \caption{The state prediction error (mean $\pm$ std, over five runs) of all models on three simpler domains. Note that models always consume predictions $\hat{s}$ in testing to match the inference procedure. The Latent-ODE achieves the lowest prediction errors on Acrobot and HIV tasks. 
    }
    \begin{tabular}{cccccccc}
    \toprule
    & RNN & $\Delta t$-RNN & Decay-RNN & Latent-RNN & ODE-RNN & Latent-ODE \\ \midrule
    Gridworld & 0.894 $\pm$ 0.023 & \textbf{0.334} $\pm$ \textbf{0.023} & 0.899 $\pm$ 0.022 & 1.161 $\pm$ 0.039 & 0.452 $\pm$ 0.040 & 0.845 $\pm$ 0.017 \\
    Acrobot & 0.176 $\pm$ 0.010 & 0.039 $\pm$ 0.006 & 0.060 $\pm$ 0.006 & 0.176 $\pm$ 0.010 & \textbf{0.022} $\pm$ \textbf{0.005} & \textbf{0.021} $\pm$ \textbf{0.005} \\
    HIV & 0.332 $\pm$ 0.013 & 0.168 $\pm$ 0.014 & 0.346 $\pm$ 0.022 & 0.361 $\pm$ 0.017 & 0.068 $\pm$ 0.006 & \textbf{0.020} $\pm$ \textbf{0.001} \\
    \bottomrule 
    \end{tabular}
    \label{tab:mse}
\end{table}
\begin{table}[t]
    \scriptsize
    \centering
    \setlength{\tabcolsep}{1.5pt}
    \vspace{-0.3cm}
    \caption{The cumulative reward (mean $\pm$ std, over five runs) of policies developed with all models on three domains. ``Oracle'' (\emph{italic}) refers to the model-free baseline trained until convergence. (a) planning in the original irregular time schedule; (b) planning using pretrained models from the original irregular time schedule for a new regular time schedule (gridworld: $\tau=7$; acrobot: $\tau=0.2$; HIV: $\tau=5$).}
    \begin{subtable}[h]{0.01\textwidth}
    \caption{}
    \label{tab:reward}
    \end{subtable}%
    \hspace{0.5cm}
    \begin{subtable}[h]{0.95\textwidth}
        \begin{tabular}{ccccccccc}
        \toprule
        & & RNN & $\Delta t$-RNN & Decay-RNN & Latent-RNN & ODE-RNN & Latent-ODE & Oracle\\ \midrule
        & Gridworld & -54.02 $\pm$ 9.24 & -45.64 $\pm$ 8.22 & -48.91 $\pm$ 8.97 & -47.92 $\pm$ 8.16 & -58.50 $\pm$ 10.46 & \bftab -34.87 $\pm$ 1.96 & \emph{-34.17 $\pm$ 1.47} \\
        & Acrobot & -179.35 $\pm$ 10.49 & -106.78 $\pm$ 10.34 & -105.23 $\pm$ 10.96 & -181.64 $\pm$ 10.92 & -65.90 $\pm$ 7.06 & \bftab -54.26 $\pm$ 4.01 & \emph{-48.67 $\pm$ 3.29} \\
        & HIV ($\times 10^7$) & 0.78 $\pm$ 0.04 & 0.75 $\pm$ 0.17 & 0.95 $\pm$ 0.21 & 0.82 $\pm$ 0.05 & 11.74 $\pm$ 1.50 & \bftab 30.32 $\pm$ 2.70 & \emph{35.22 $\pm$ 1.42} \\  
        \bottomrule 
        \end{tabular}
    \end{subtable}%
    \vspace{0.3cm}
    \begin{subtable}[h]{0.01\textwidth}
    \caption{}
    \label{tab:transfer-reward}
    \end{subtable}%
    \hspace{0.5cm}
    \begin{subtable}[h]{0.95\textwidth}
        \begin{tabular}{ccccccccc}
        \toprule
        & & RNN & $\Delta t$-RNN & Decay-RNN & Latent-RNN & ODE-RNN & Latent-ODE & Oracle\\ \midrule
        & Gridworld & -61.01 $\pm$ 10.03 & -64.55 $\pm$ 10.89 & -60.78 $\pm$ 10.03 & -52.32 $\pm$ 8.91 & -114.70 $\pm$ 11.65 & \bftab -49.31 $\pm$ 6.62 & \emph{-35.93 $\pm$ 1.95} \\
        & Acrobot & -407.46 $\pm$ 13.82 & -281.92 $\pm$ 9.99 & -285.07 $\pm$ 8.47 & -237.25 $\pm$ 10.29 & -190.82 $\pm$ 9.13 & \bftab -171.37 $\pm$ 10.07 & \emph{-78.75 $\pm$ 3.23}\\
        & HIV ($\times 10^7$) & 7.66 $\pm$ 1.79 & 17.21 $\pm$ 2.44 & 5.84 $\pm$ 1.62 & 16.95 $\pm$ 3.05 & 11.32 $\pm$ 1.09 & \bftab 21.60 $\pm$ 2.39 & \emph{33.55 $\pm$ 1.97} \\  
        \bottomrule 
        \end{tabular}
    \end{subtable}
\end{table}

\paragraph{Latent-ODEs mimic continuous-time environments and value functions more accurately.} Because the training dataset is fixed, we can use the \emph{state prediction error} (MSE in Equation \ref{eq:loss-state}) on a separate test dataset to measure if the model learns the dynamics well. Table \ref{tab:mse} shows state prediction errors of all models. The ODE-RNN and Latent-ODE outperform other models on acrobot and HIV, but the $\Delta t$-RNN achieves the lowest error on the windy gridworld. However, by visualizing the value functions of learned policies which are constructed using dynamics models (Figure \ref{fig:value-func}), we find that only the Latent-ODE accurately recovers the true gridworld (the one from the model-free baseline), whereas the $\Delta t$-RNN characterizes the dark parts of the world very well, but fails to identify the true goal region (the light part). Thus, the lower state prediction error averaged over the gridworld does not imply better policies.
 
\paragraph{Latent-ODEs help agents develop better policies.} 
Table \ref{tab:reward} shows the performance of all models, in terms of returns achieved by policies in the actual environment. Latent-ODE consistently surpasses other models and achieves near-optimal performance with the model-free baseline. In contrast, all non-ODE-based models develop very poor policies on the acrobot and HIV problems.

\paragraph{Latent-ODEs are more robust to changes in time intervals.} 
To test if the dynamics model is generalizable across interval times, we change the time schedules from irregular measurements to regular measurements without retraining the models. Due to the page limit, we show the results of $\tau=7$ for the gridworld, $\tau=0.2$ for the acrobot and $\tau=5$ for HIV in Table \ref{tab:transfer-reward} and include full results of all time discretizations in Appendix \ref{apx:res-regular}. The Latent-ODE is once again the best dynamics model to solve the new environment, even if the transition times are regular.

\paragraph{Optimized time schedules achieve the best balance of high performance and low interaction rate.}
\begin{wrapfigure}{r}{0.47\textwidth}
\centering
\vspace{-0.4cm}
\includegraphics[width=0.47\textwidth]{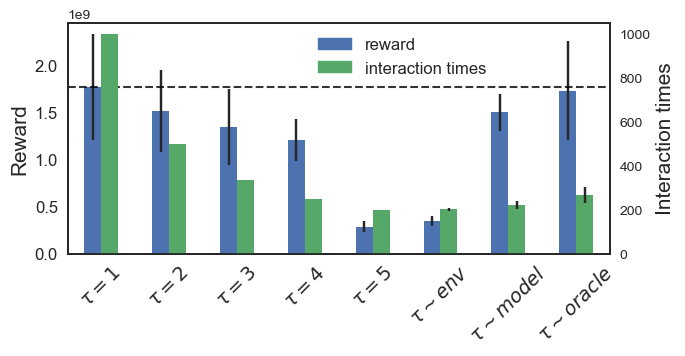}
\vspace{-0.65cm}
\caption{The cumulative reward vs. interaction rate on the HIV environment. ``$\tau \sim$ env'' refers to the original time schedule for training models; ``$\tau \sim$ model/oracle'' refers to selecting $\tau$ using the Latent-ODE/true dynamics and Equation \ref{eq:opt}.
}
\vspace{-0.5cm}
\label{fig:opt}
\end{wrapfigure}
For evaluation, to ensure the fair comparison of different interaction rates given the fixed horizon, we collect the reward at every time step (every day) from the environment and calculate the overall cumulative reward. The results on the HIV environment are shown in Figure \ref{fig:opt}. 
Developing the model-based schedule using the Latent-ODE, we can obtain similar returns as measuring the health state every two days, but with less than half the interventions. Further, using the oracle dynamics, the optimized schedule achieves similar performance with maximal interaction ($\tau=1$) while reducing interaction times by three-quarters. 

\begin{wraptable}{r}{3.7cm}
\scriptsize
\vspace{-0.45cm}
\caption{The performance of different policies on partially observable HIV task.
}
\label{tab:hiv-pomdp}
\vspace{-0.45cm}
\setlength{\tabcolsep}{0.5pt}
\begin{tabular}{cc}\\\toprule  
Policy & Reward $(\times 10^7)$ \\\midrule
$\pi^{MB}(a|s_\text{partial})$ & 12.88 $\pm$ 1.71\\ 
$\pi^{MB}(a|s_\text{partial}, z)$ & \bftab 18.04 $\pm$ 2.07 \\
$\pi^{MF}(a|s_\text{partial})$ & \emph{16.96 $\pm$ 1.53} \\ $\pi^{MB}(a|s_\text{full})$ & \emph{30.32 $\pm$ 2.70} \\
\bottomrule
\end{tabular}
\vspace{-0.3cm}
\end{wraptable} 
\paragraph{Latent variables capture hidden state representations in partially observable environment.}
We mask two blood test markers in the state space to build a partially observable HIV environment, where we demonstrate the behavior of the latent policy $\pi^{MB}(a|s_\text{partial}, z)$. We compare its performance with the model-based policy $\pi^{MB}(a|s_\text{partial})$ and the model-free policy $\pi^{MF}(a|s_\text{partial})$ using partial observations, and the model-based policy $\pi^{MB}(a|s_\text{full})$ using full observations. All model-based policies are developed with the Latent-ODE. Based on the results in Table \ref{tab:hiv-pomdp}, the latent variable $z$ improves the performance in the partially observable setting, and even builds a better policy than the model-free baseline, though the latent policy cannot achieve the asymptotic performance of the policy using full observations.

\begin{figure}[t]
    \centering
    \includegraphics[width=\textwidth]{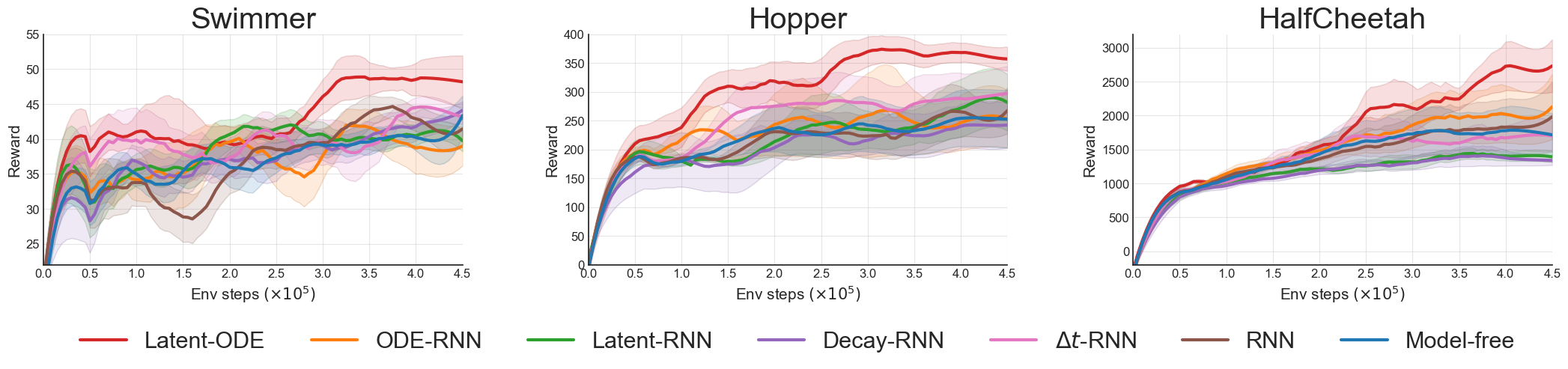}
    \caption{Learning curves with all models on three Mujoco tasks. The shaded region represents a standard deviation of average evaluation over four runs (evaluation data is collected every 5000 timesteps). Curves are smoothed with a 20-point window for visual clarity. The Latent-ODE develop better policies with fewer data than the model-free method and other models on all three tasks.} 
    \label{fig:reward-mujoco}
\end{figure}

\subsection{Continuous control: Mujoco}
For the more complex Mujoco tasks, exploration and learning must be interleaved. We combine model predictive control (MPC) \citep{mayne2000constrained} with the actor-critic method (DDPG) for planning. MPC refines the stochastic model-free policy (the actor) by sampling-based planning \citep{wang2019exploring, hong2019model}, and the value function (the critic) mitigates the short sight of imaginary model rollouts in MPC planning \citep{lowrey2018plan, clavera2020model}. This approach iterates between data collection, model training, and policy optimization, which allows us to learn a good policy with a small amount of data. The details of the algorithm can be found in Appendix \ref{apx:algo-mpc-ac}. 

\paragraph{Latent-ODEs exhibit the sample efficiency across all three Mujoco tasks.} 
Figure \ref{fig:reward-mujoco} shows the learning process of all models on Mujoco tasks. The Latent-ODE is more sample-efficient than the model-free method and other models on all three tasks. For example, on the swimmer and hopper task, we develop a high-performing policy over 100k environment steps using the Latent-ODE, whereas the model-free baseline requires four times the amount of data. However, the ODE-RNN is not as good as the Latent-ODE and its performance is similar with other baselines.

\section{Discussion and conclusion}
We incorporate actions and time intervals into neural ODEs for modeling continuous-time dynamics, and build a unified framework to train dynamics models for SMDPs. Our empirical evaluation across various continuous-time domains demonstrates the superiority of the Latent-ODE in terms of model learning, planning, and robustness for changes in interval times. Moreover, we propose a method to minimize interventions with the environment but maintain near-optimal performance using the dynamics model, and show its effectiveness in the health-related domain.

The flexibility of our model training procedure (Section \ref{sec:model}), which is orthogonal to the planning method, inspires the future research on model-based RL and neural ODEs in various aspects. First, one might easily enhance the performance of Latent-ODEs in continuous-time domains using more advanced planning schemes and controllers, e.g., TD3 \citep{fujimoto2018addressing}, soft actor-critic \citep{haarnoja2018soft} and etc. Second, since our model is always trained on a batch of transition data, we can apply our method to the continuous-time off-policy setting with the recent advances on model-based offline RL \citep{yu2020mopo, kidambi2020morel}. RL health applications (e.g., in ICU \citep{gottesman2020interpretable}) might benefit from this in particular, because practitioners usually assume the presence of a large batch of already-collected data, which consists of patient measurements with irregular observations and actions. Furthermore, while we focus on flat actions in this work, it is natural to extend our models and framework to model-based hierarchical RL \citep{botvinick2014model, li2017efficient}. 

Last but not least, we find that training a Latent-ODE usually takes more than ten times longer than training a simple RNN model due to slow ODE solvers, which means the scalability might be a key limitation for applying our models to a larger state space setting. We believe that the efficiency of our methods will not only be significantly improved with a theoretically faster numerical ODE solver (e.g., \citep{finlay2020train, kelly2020learning}), but also with a better implementation of ODE solvers\footnote{We use the implementation of ODE solvers from Python \href{https://github.com/rtqichen/torchdiffeq}{torchdiffeq} library.} (e.g., a faster C\texttt{++}/Cython implementation, using single precision arithmetic for solvers, and etc.).

\section*{Broader Impact}
We introduce a new approach for continuous-time reinforcement learning that could eventually be useful for a variety of applications with irregular time-series, e.g. in healthcare. However, models are only as good as the assumptions made in the architecture, the data they are trained on, and how they are integrated into a broader context.  Practitioners should treat any output from RL models objectively and carefully, as in real life there are many novel situations that may not be covered by the RL algorithm. 

\section*{Acknowledgement}
We thank Andrew Ross, Weiwei Pan, Melanie Pradier and other members from Harvard Data to Actionable Knowledge lab for helpful discussion and feedbacks. We thank Harvard Faculty of Arts and Sciences Research Computing and School of Engineering and Applied Sciences for providing computational resources.  FDV and JD are supported by an NSF CAREER.

\bibliographystyle{abbrvnat}
\bibliography{bibliography}

\clearpage
\begin{center}
\huge
Supplementary Materials
\end{center}

\begin{appendices}

\section{Algorithm details}\label{apx:algo}
\subsection{Optimizing interval times}\label{apx:algo-opt}

\begin{algorithm}[htb]
    \SetKwInOut{Input}{Input}\SetKwInOut{Output}{Output}
    \Input{The pretrained state transition dynamics model $\hat{P}(z'|z,a,s,\tau)$, the replay buffer $\mathcal{D}$, the reward function $R(s,a,s')$, the time horizon $T$, and the number of episodes $N_e$.}
    \Output{$\pi_\omega(a|s)$.}
    Initialize the policy $\pi_\omega(a|s)$ (along with the action-value function $Q^\pi(s,a)$);\\
    \For{$j \leftarrow 1$ \KwTo $N_e$}{
        $t \leftarrow 0$;\\
        Observe the initial state $s$ from the environment;\\
        Initialize the latent state $z$;\\
        \While{$t < T$}{
            Select action $a \sim \pi_\omega(\cdot|s)$;
            \textcolor{blue}{\\$\hat{s}_{0} \leftarrow s$;\\
            \For{$i \leftarrow 1$ \KwTo $|\mathcal{T}|$}{
                $z_i \leftarrow \hat{P}(\cdot|z,a,s,i)$;\\
                Decode $\hat{s}_i$ from $z_i$;\\
                Select action $\hat{a}_i \sim \pi_\omega(\cdot|\hat{s}_i)$;\\
                Calculate the immediate reward $\hat{r}_i \leftarrow R(\hat{s}_{i-1}, a, \hat{s}_i)$ at the current time point $t + i$;\\
            }
            Select the best incoming time interval $\tau^\ast \leftarrow \argmax_\tau \sum_{i=1}^{\tau} \gamma^{i-1} \hat{r}_i + \gamma^\tau Q^\pi(\hat{s}_\tau, \hat{a}_\tau)$;\\
            $z', s', r \leftarrow z_{\tau^\ast}, \dot{s}_{\tau^\ast}, \sum_{i=1}^{\tau^\ast} \gamma^{i-1} \dot{r}_i$;\\}
            \eIf{$s'$ is not the terminated state}{
                Store the tuple $(s,a,s',r,\tau)$ into $\mathcal{D}$;\\
            }{
                Store the tuple $(s,a,\text{NULL},r,\tau)$ into $\mathcal{D}$;\\
                \textbf{break};\\
            }
            Optimize $\pi_\omega$ with data in $\mathcal{D}$;\\
            $t, s, z \leftarrow t + \tau^\ast, s', z'$;\\
        }
    }
    \caption{Optimizing interval times with the dynamics model.}
    \label{algo:opt}
\end{algorithm}
Our innovation of optimizing interval times is highlighted in \textcolor{blue}{blue} in Algorithm \ref{algo:opt}. Note that we can use either the imaginary reward $\hat{r}$ from the dynamics model or the true reward $r$ from the environment for training the policy $\pi_\omega$, i.e., $\dot{r}$ can be either $\hat{r}$ or $r$, and similar case for $\dot{s}_{\tau^\ast}$ (line 16 of Algorithm \ref{algo:opt}). In this work, to focus on the efficacy of the optimized time schedules, we use the true reward $r$ and true observation $s_{\tau^\ast}$; but the optimal $\tau$ is always determined using the imaginary reward $\hat{r}$. 

A key assumption of Algorithm \ref{algo:opt} is that acting often using short time intervals will not hurt performance, and that maximal interaction (i.e. $\tau=1$ for discrete time) will have the optimal performance. In many scenarios, this assumption seems reasonable and applying Algorithm \ref{algo:opt} may work well.  For instance, in healthcare applications, it is safe to assume that more frequent monitoring of patients and more careful tuning of their treatment plan should yield optimal performance, although in practice this is not generally done due to e.g. resource constraints. However, this assumption does not hold for all environments. For example, some Atari games require frameskipping, i.e., repeating actions for $k$ times, because we need enough changes in pixels to find a good policy. Also, some control problems have an optimal time step for the underlying physical systems. In these situations, one might set a minimum threshold $a$ for $\tau$ and apply the algorithm, i.e., $\tau \in [a, +\infty)$; together with the extension to continuous time (Equation \ref{eq:opt-cont}), we leave them to future work. 

\begin{gather}
    \tau^\ast := \argmax_{\tau \in [a, \infty)} \int_{t=a}^{\tau} \gamma^{t} \hat{r} \; dt + \gamma^\tau Q^\pi(\hat{s}_\tau, \hat{a}_\tau) 
    \label{eq:opt-cont}
\end{gather}

\subsection{Learning world models}\label{apx:algo-world-model}
\begin{algorithm}[htb]
    \SetKwInOut{Input}{Input}\SetKwInOut{Output}{Output}
    \Input{The replay buffer $\mathcal{D}$, the reward function $R(s,a,s')$, the time horizon $T$, and the number of episodes $N_t$ for model learning and $N_e$ for policy optimization.}
    \Output{$\hat{P}(z',\tau|z,a,s), \pi_\omega(a|s)$.}
    Initialize the policy $\pi_\omega(a|s)$ (along with the action-value function $Q^\pi(s,a)$);\\
    Initialize the dynamics model $\hat{P}(z',\tau|z,a,s)$;\\
    Collect a collection of trajectories $h=\left(\{s^{(n)}_i\}_{i=0}^{L_n},\{a^{(n)}_i\}_{i=0}^{L_n-1},\{\tau^{(n)}_i\}_{i=0}^{L_n-1} \right)_{n=1}^{N_t}$ using random policies, where $\sum_{i=0}^{L_n-1} \tau^{(n)}_i < T$ for $\forall n\in\{1, 2, \ldots, N_t\}$; \\
    Train $\hat{P}(z', \tau|z,a,s)$ with $h$ as described in Section \ref{sec:model}; \\
    \For{$i \leftarrow 1$ \KwTo $N_e$}{
        $t \leftarrow 0$;\\
        Observe the initial state $s$ from the environment or sample from a set of initial states;\\
        Initialize the latent state $z$;\\
        \While{$t < T$}{
            Select action $a \sim \pi_\omega(\cdot|s)$;\\
            Predict the incoming time interval $\tau$ and next latent state $z'$ using $\hat{P}(\cdot|z,a,s)$;\\
            Decode $s'$ from $z'$;\\
            Calculate the reward $r \leftarrow R(s,a,s')$;\\
            \eIf{$s'$ is not the terminated state}{
                Store the tuple $(s,a,s',r,\tau)$ into $\mathcal{D}$;\\
            }{
                Store the tuple $(s,a,\text{NULL},r,\tau)$ into $\mathcal{D}$;\\
                \textbf{break};\\
            }
            Optimize $\pi_\omega$ with data in $\mathcal{D}$;\\
            $t, s, z \leftarrow t + \tau, s', z'$;\\
        }
    }
    \caption{Learning world models \citep{ha2018world} for SMDPs.}
    \label{algo:world-model}
\end{algorithm}

Algorithm \ref{algo:world-model} assumes that the dynamics can be fully covered by random policies. However, these may be far away from the optimal policy. Because the policy is trained only on fictional samples without considering the planning horizon, there is no difference between learning in the virtual world created by the dynamics model and the true environment. Thus, the performance of learned policies is mainly determined by the model's capacity.

Nevertheless, Algorithm \ref{algo:world-model} does not work well on more sophisticated Mujoco tasks, because the dynamics cannot be fully explored by a random policy, and a long planning horizon makes the compounding error of fictional samples accumulate very quickly. One might use an iterative training procedure of Algorithm \ref{algo:world-model} \citep{ha2018world, schmidhuber2015learning} for more complex environments, which interleaves exploration and learning. However, in order to combat the model bias, this type of Dyna-style algorithm usually requires computationally expensive model ensembles \citep{kurutach2018model, janner2019trust}. Thus, we turn to the MPC-style algorithm for planning (Algorithm \ref{algo:mpc-ac}), which is also sufficiently effective to demonstrate a learned model's capacity and is more computationally efficient. 

\clearpage
\subsection{Model predictive control with actor-critic}\label{apx:algo-mpc-ac}

\begin{algorithm}[htb]
    \SetKwInOut{Input}{Input}\SetKwInOut{Output}{Output}
    \Input{The replay buffer $\mathcal{D}$, the environment dataset $\mathcal{D}_{env}$, the reward function $R(s,a,s')$, the planning horizon $H$, the search population $K$, the number of environment steps $M$ and the number of epochs $E$.}
    \Output{$\hat{P}(z', \tau|z,a,s), \pi_\omega(a|s), Q^\pi_ \varphi(s,a)$.}
    Initialize the actor $\pi_\omega(a|s)$ and the critic $Q^\pi_\varphi(s,a)$, and their target networks $\pi_{\omega'}$ and $Q^\pi_{\varphi'}$;\\
    Initialize the dynamics model $\hat{P}(z',\tau|z,a,s)$;\\
    Gather a collection of trajectories using random policies, and save them into $\mathcal{D}_{env}$;\\
    \For{$i \leftarrow 1$ \KwTo $E$}{
        Train $\hat{P}(z',\tau|z,a,s)$ with data in $\mathcal{D}_{env}$ as described in Section \ref{sec:model}; \\
        Observe the initial state $s$ from the environment;\\
        Initialize the latent state $z$;\\
        \For{$j \leftarrow 1$ \KwTo $M$}{
            \For{$k \leftarrow 1$ \KwTo $K$}{
                $t^{(k)}_0, \hat{s}^{(k)}_0, z^{(k)}_0 \leftarrow 0, s, z$;\\
                \For{$h \leftarrow 1$ \KwTo $H$}{
                    Select action $a^{(k)}_{h-1} \sim \pi_\omega(\cdot|\hat{s}^{(k)}_{h-1})$;\\
                    $z^{(k)}_h, \tau^{(k)}_{h-1} \leftarrow \hat{P}(\cdot|z^{(k)}_{h-1},a^{(k)}_{h-1},\hat{s}^{(k)}_{h-1})$;\\
                    Decode $\hat{s}^{(k)}_h$ from $z^{(k)}_h$;\\
                    Calculate the reward $\hat{r}^{(k)}_{h-1} \leftarrow R(\hat{s}^{(k)}_{h-1},a^{(k)}_{h-1},\hat{s}^{(k)}_h)$;\\
                    $t^{(k)}_h \leftarrow t^{(k)}_{h-1} + \tau^{(k)}_{h-1}$;\\
                }
                Select action $a^{(k)}_H \leftarrow \pi_\omega(\cdot|\hat{s}^{(k)}_H)$;\\
            }
            Select the best sequence index $k^\ast \leftarrow \argmax_{k} \sum_{h=1}^H \gamma^{t^{(k)}_{h-1}} \hat{r}^{(k)}_{h-1} + \gamma^{t^{(k)}_H} Q^\pi_\varphi(\hat{s}^{(k)}_H, a^{(k)}_H)$;\\
            Select the best action $a \leftarrow a^{(k^\ast)}_0$;\\
            Observe the incoming time interval $\tau$ and next observation $s'$;\\
            Encode the next latent state $z' \leftarrow \hat{P}(\cdot|z,a,s,\tau)$;\\
            Calculate the reward $r \leftarrow R(s,a,s')$;\\
            \eIf{$s'$ is not the terminated state}{
                Store the tuple $(s,a,s',r,\tau)$ into $\mathcal{D}$;\\
            }{
                Store the tuple $(s,a,\text{NULL},r,\tau)$ into $\mathcal{D}$;\\
                Observe the initial state $s$ from the environment;\\
                Initialize the latent state $z$;\\
                \textbf{continue};\\
            }
            Optimize $\pi_\omega$ and $Q^\pi_\varphi$ with data in $\mathcal{D}$;\\
            Update target networks $\pi_{\omega'}$ and $Q^\pi_{\varphi'}$;\\
            $s, z \leftarrow s', z'$;\\
        }
        Store trajectories collected in the current epoch into $\mathcal{D}_{env}$;\\
    }
    \caption{Model predictive control (MPC) with DDPG for SMDPs.}
    \label{algo:mpc-ac}
\end{algorithm}

In Algorithm \ref{algo:mpc-ac}, the actor provides a good initialization of deterministic action sequences, and MPC searches for the best one from these sequences plus Gaussian noises, i.e., we sample the action from a normal distribution $a \sim \mathcal{N}(\pi_\omega(\cdot|s), \sigma^2)$, where $\sigma^2$ is the noise variance (line 12 of Algorithm \ref{algo:mpc-ac}). Further, in the vanilla MPC, the best sequence is determined by the cumulative reward of model rollouts, however, the selected action may not be globally optimal due to the short planning horizon $H$. The critic offers an estimate of expected returns for the final state of simulated trajectories, which overcomes the shortsighted planning problem (line 20 of Algorithm \ref{algo:mpc-ac}). Note that we use the deterministic action $a_H$ (without Gaussian noise) for the final state $\hat{s}_H$ to calculate the action-value function $Q(\hat{s}_H, a_H)$ (line 18 of Algorithm \ref{algo:mpc-ac}).

\section{Environment specifications}\label{apx:env}
\paragraph{Windy gridworld.}
We extend the $7 \times 10$ gridworld to have continuous states $s = (x, y)$ and continuous-time actions. That is, picking an orientation (action) $a=(\Delta x, \Delta y)$, agents move towards this direction over arbitrary second(s) $\tau \in \text{Unif}\{1,7\}$ for $(\tau\Delta x, \tau\Delta y)$. The agent is allowed to ``move as a king'', i.e., take eight actions, including moving up, down, left, right, upleft, upright, downleft and downright. Specifically, the agent can move $a = (\Delta x, \Delta y)$ in the gridworld every second:
\begin{gather*}
    a = (\Delta x, \Delta y) = 
    \begin{cases}
        (0.17, 0) & \text{ up }, \\
        (-0.17, 0) & \text{ down }, \\
        (0, -0.17) & \text{ left }, \\
        (0, 0.17) & \text{ right }, \\
        (\frac{0.17}{\sqrt{2}}, -\frac{0.17}{\sqrt{2}}) & \text{ upleft }, \\
        (\frac{0.17}{\sqrt{2}}, \frac{0.17}{\sqrt{2}}) & \text{ upright }, \\
        (-\frac{0.17}{\sqrt{2}}, -\frac{0.17}{\sqrt{2}}) & \text{ downleft }, \\
        (-\frac{0.17}{\sqrt{2}}, \frac{0.17}{\sqrt{2}}) & \text{ downright }.
    \end{cases}
\end{gather*}
Every second, agents in the region with the weaker wind $(x \in [2.5, 8.5], y \in [0, 7])$ are pushed to move upward for $(\Delta x, \Delta y) = (\frac{0.17}{4}, 0)$, and agents in the region with the strong wind $(x \in [5.5, 7.5], y \in [0, 7])$ are pushed to move upward for $(\Delta x, \Delta y) = (\frac{0.17}{2}, 0)$. If agents hit the boundary of the gridworld, they will just stand still until the end of transition time $\tau$. Every second in the gridworld incurs -1 cost until discovering the goal region (trigger +10 reward) or after $T=150$ seconds. Thus, we are given the reward function
\begin{gather*}
    R(s,a,s',\tau) = 
    \begin{cases}
    10-\tau, & s' = (x \in [3, 4], y \in [6.5, 7.5]),\\
    -\tau, & \text{otherwise.}
    \end{cases}
\end{gather*}
In addition, we feed the zero-centered state $(\bar{x},\bar{y})$ for both model training and policy optimization.

\paragraph{Acrobot.} 
Acrobot, a canonical RL and control problem, is a two-link pendulum with only the second joint actuated. Initially, both links point downwards. The goal is to swing up the pendulum by applying a positive, neutral, or negative torque on the joint such that the tip of the pendulum reaches a given height. The state space consists of four continuous variables, $s = (\theta_1, \theta_2, \dot{\theta}_1, \dot{\theta}_2)$, where $\theta_1 \in [-\pi, \pi]$ is the angular position of the first link in relation to the joint, and $\theta_2\in [-\pi, \pi]$ is the angular position of the second link in relation to the first; $\dot{\theta}_1 \in [-4\pi, 4\pi]$ and $\dot{\theta}_2 \in [-9\pi, 9\pi]$ are the angular velocities of each link respectively. The reward is collected as the height of the tip of the pendulum (as recommended in the work of \citep{wang2019benchmarking}) after the transition time $\tau$:
\begin{gather*}
    R(s,a,s') = -\cos{\theta_1} - \cos{(\theta_1 + \theta_2)},
\end{gather*}
until the goal is reached or after $T=100$.

\paragraph{HIV.} The interaction of the immune system with the human immunodeficiency virus (HIV) and treatment protocols was mathematically formulated as a dynamical system \citep{adams2004dynamic}, which can be resolved using RL approaches \citep{ernst2006clinical, parbhoo2017combining, killian2017robust}. The goal of this task is to determine effective treatment strategies for HIV infected patients based on critical markers from a blood test, including the viral load ($V$, which is the main maker indicating if healthy), the number of healthy and infected CD4+ T-lymphocytes ($T_1$, $T_1^\ast$, respectively), the number of healthy and infected macrophages ($T_2$, $T_2^\ast$, respectively), and the number of HIV-specific cytotoxic T-cells ($E$), i.e., $s = (T_1, T_2, T_1^\ast, T_2^\ast, V, E)$. To build a partially observerable HIV environment, $T_1^\ast$ and $T_2^\ast$ are removed from the state space. The anti-HIV drugs can be roughly grouped into two main categories (Reverse Transcriptase Inhibitors (RTI) and Protease Inhibitors (PI)). The patient is assumed to be given treatment from one of two classes of drugs, a mixture of the two treatments, or provided no treatment. The agent starts at an unhealthy status $s_0 = [163573, 5, 11945, 46, 63919, 24]$, where the viral load and number of infected cells are much higher than the number of virus-fighting T-cells. The dynamics system is defined by a set of differential equations:
\begin{equation*}
    \begin{aligned} 
        \dot{T}_{1} &=\lambda_{1}-d_{1} T_{1}-\left(1-\epsilon_{1}\right) k_{1} V T_{1} \\ \dot{T}_{2} &=\lambda_{2}-d_{2} T_{2}-\left(1-f \epsilon_{1}\right) k_{2} V T_{2} \\ \dot{T}_{1}^{*} &=\left(1-\epsilon_{1}\right) k_{1} V T_{1}-\delta T_{1}^{*}-m_{1} E T_{2}^{*} \\ \dot{T}_{2}^{*} &=\left(1-\epsilon_{2}\right) N_{T} \delta\left(T_{1}^{*}+T_{2}^{*}\right)-c V-\left[\left(1-\epsilon_{1}\right) \rho_{1} k_{1} T_{1}+\left(1-f \epsilon_{1}\right) \rho_{2} k_{2} T_{2}\right] V \\ \dot{E} &=\lambda_{E}+\frac{b_{E}\left(T_{1}^{*}+T_{2}^{*}\right)}{\left(T_{1}^{*}+T_{2}^{*}\right)+K_{b}} E-\frac{d_{E}\left(T_{1}^{*}+T_{2}^{*}\right)}{\left(T_{1}^{*}+T_{2}^{*}\right)+K_{d}} E-\delta_{E} E 
    \end{aligned}
\end{equation*}
where $\epsilon_1=0.7$ (if RTI is applied, otherwise 0) and $\epsilon_2=0.3$ (if PI is applied, otherwise 0) are treatment specific parameters, selected by the prescribed action. See the specification of other parameters in the work of \citep{adams2004dynamic}.

The effective period is determined by the state (mainly determined by the viral load $V$) and the treatment as follow:
\begin{gather*}
    \tau \sim 
    \begin{cases}
        \text{Unif}\{7,14\} &\text{ if } V \leq 10^4 \text{ and no treatment, } \\
        \text{Unif}\{3,7\} &\text{ if } V \leq 10^4 \text{ and any treatment, } \\
        \text{Unif}\{3,7\} &\text{ if } 10^4 \leq V \leq 10^5 \text{ and no treatment, } \\
        \text{Unif}\{3,5\} &\text{ if } 10^4 \leq V \leq 10^5 \text{ and (only RIT or only PI), } \\
        3 &\text{ if } 10^4 \leq V \leq 10^5 \text{ and (both RIT and PI), } \\
        3 &\text{ if } V \geq 10^5 \text{ and no treatment, } \\
        \text{Unif}\{1,2\} &\text{ if } V \geq 10^5 \text{ and any treatment. }
    \end{cases}
\end{gather*}
The reward is gathered based on the patient's healthy state after effective period:
\begin{gather*}
    R(s,a,s')= -0.1V - 20000\epsilon_1^2 - 2000\epsilon_2^2 + 1000E.
\end{gather*}
An episode ends after $T=1000$ days and there is no early terminated condition. The state variables are first put in log scale then normalized to have zero mean and unit standard deviation for both model training and policy optimization. 

\paragraph{Mujoco.} 
We consider the fully observable Mujoco environments, where the position of the root joint is also observed, which allows us to calculate the reward and determine the terminal condition for simulated states easily. The action repeats $\tau$ times with the following pattern:
\begin{gather*}
    \tau = \left\lfloor\frac{d-c}{2}\cos{\left(20\pi\|\theta_v\|_2\right)} + \frac{c+d}{2}\right\rceil
\end{gather*}
where $c/d$ is the minimum/maximum action repetition times, $\theta_v$ is the angle velocity vector of all joints and $\lfloor\cdot\rceil$ is rounding to the nearest integer. We have $c=1$ for all three locomotion tasks; $d=7$ for the swimmer and the hopper and $d=9$ for the half-cheetah. 

Because we assume the intermediate observations are not available during action repetition, the reward is calculated only based on the current observation $s$, the next observation $s'$, the control input $a$ and and repeated times $\tau$:
\begin{gather*}
    R(s,a,s',\tau) = 
    \begin{cases}
    \frac{x'-x}{\tau} - 0.0001\|a\|^2_2, & \; \text{for the swimmer,}\\
    \frac{x'-x}{\tau} - 0.001\|a\|^2_2 + \mathbbm{1}(\text{alive}), & \; \text{for the hopper,}\\
    \frac{x'-x}{\tau} - 0.1\|a\|^2_2, & \; \text{for the half-cheetah,}
    \end{cases}
\end{gather*}
where $x/x'$ is the previous/current position of the root joint, and there is an alive bonus of 1 for the hopper for every step. Also, instead of setting a fixed horizon, we keep the original maximum length of an episode in OpenAI Gym, i.e., the maximum number of environment steps over an episode is 1000 for all three tasks. We normalize observations so that they have zero mean and unit standard deviation for both model training and policy optimization.

\section{Experimental details}
\subsection{Planning}
\paragraph{Learning world models.} 
For all three simpler domains, we collect $N_t=1000$ episodes as the training dataset, and collect another 100 episodes as the validation dataset. We optimize the policy for $N_e$ episodes until convergence, whose value is shown in Table \ref{tab:ne}. All final cumulative rewards are evaluated by taking the average reward of 100 trials after training policies for $N_e$ episodes. 

\begin{table}[htb]
\centering
\caption{The choice of $N_e$ for training policies until convergence.}
\begin{tabular}{|c|c|c|c|}
\hline
& Gridworld & Acrobot & HIV \\ \hline
model-based & 1000 & 200 & 1500 \\ \hline
model-free & \multicolumn{2}{c|}{1000} & 3500 \\ \hline
\end{tabular}
\label{tab:ne}
\end{table}

\paragraph{Model predictive control with actor-critic.}
We switch between model training and policy optimization every $M=5000$ environment steps. Equipped with the value function from the critic, we can choose a relatively shorter planning horizon $H=10$, which maintains the good performance while reducing the computational cost. We set a large search population $K=1000$. 

For model learning, the 90\% collected trajectories are used as the training dataset and the remaining 10\% are used as the validation dataset. Also, we divide a full trajectory into several pieces, whose length is equal to (or less than) the MPC planning horizon $H$. Not only does it reduce the computational cost of training a sequential model, but also helps the dynamics model provide more accurate predictions in a finite horizon. 

Throughout all experiments, we use a soft-greedy trick for MPC planning to combat the model approximation errors \citep{hong2019model}. Instead of selecting the best first action (line 20 of Algorithm \ref{algo:mpc-ac}), we take the average of first actions of the top 50 sequences as the final action for the agent. This simple approach alleviates the impact of inaccurate models and improves the performance.

\paragraph{Initialization of latent states.} 
While using the model for planning, initial latent states of recurrent-based models are all zeros, but they are sampled from the prior distribution (standard normal distribution) for models with the encoder-decoder structure.

\subsection{Model learning}\label{apx:train}
\paragraph{Scheduled sampling.} 
As our model predicts the new latent state $z'$ at time $t+1$, it needs to be conditioned on the previous state at the previous time step $t$. During training, there is a choice for the source of the next input for the model: either the ground truth (observation) or the model’s own previous prediction can be taken. The former provides more signal when the model is weak, while the latter matches more accurately the conditions during inference (episode generation), when the ground truth is not known. The scheduled sampling strikes a balance between the two. Specifically, at the beginning of the training, the ground truth is offered more often, which pushes the model to deliver the accurate short term predictions; at the end of the training, the previous predicted state is more likely to be used to help the model focus on the global dynamics. In other word, the optimization objective transits from the one-step loss to multiple-step loss. Therefore, the scheduled sampling can prevent the model from drifting out of the area of its applicability due to compounding errors. We use a linear decay scheme for scheduled sampling $\epsilon = \max\{0, 1 - c \cdot k\}$, where $c = 0.0001$ is the decay rate and $k$ is the number of iterations. However, we find that the scheduled sampling only works well on the acrobot task.  

\paragraph{Early stopping.} On Mujoco tasks, we utilize early stopping to prevent from overfitting, i.e., we terminate the training if the state prediction error (MSE in Equation \ref{eq:loss-state}) on the validation dataset does not decrease for $e$ training epochs and we use the parameters achieving the lowest state prediction error as the final model parameters. Because the model already learns the dynamics after trained for several epochs in Algorithm \ref{algo:mpc-ac} and only needs to be refined for some novel situations, we use a linear decay scheme: $e=\max(15-k, 3)$, where $k$ is the number of epochs in Algorithm \ref{algo:mpc-ac}. For three simpler domains, we run 12,000 iterations without early stopping. 

\paragraph{Model hyperparameters.}
We tune the hyperparameters for dynamics models on different domains, but all baseline models use a same set of hyperparameters for comparison. 

\begin{table}[htb]
\centering
\caption{Hyperparameters for dynamics models on different domains.}
\begin{tabular}{|c|c|c|c|c|c|c|}
\hline
& Gridworld & Acrobot & HIV & Swimmer & Hopper & HalfCheetah \\ \hline
Learning rate & 1e-3 & \multicolumn{1}{|c|}{5e-4} & \multicolumn{4}{c|}{1e-3} \\ \hline
Batch size & \multicolumn{3}{c}{32} & \multicolumn{3}{|c|}{128} \\ \hline
Latent dimension & 2 & \multicolumn{2}{|c|}{10} & \multicolumn{2}{c|}{128} & 400\\ \hline
Weight decay & \multicolumn{6}{c|}{1e-3}\\ \hline
Scheduled sampling & No & Yes & \multicolumn{4}{c|}{No}\\ \hline
GRU & \multicolumn{6}{c|}{one layer, unidirectional, Tanh activation}\\ \hline
Encoder hidden to latent & \multicolumn{1}{c|}{5} & \multicolumn{5}{c|}{20} \\ \hline
Interval timer $g_\kappa$ & \multicolumn{2}{c|}{N/A} & \multicolumn{4}{c|}{20}\\ \hline
$\lambda$ in Equation \ref{eq:loss} & \multicolumn{2}{c|}{0} & \multicolumn{4}{c|}{0.01}\\ \hline
ODE network $f_\theta$ & \multicolumn{1}{c|}{5} & \multicolumn{2}{c|}{20} & \multicolumn{2}{c|}{128} & \multicolumn{1}{c|}{400} \\ \hline
ODE solver & \multicolumn{6}{c|}{Runge-Kutta 4(5) adaptive solver (dopri5)} \\ \hline
ODE error tolerance & \multicolumn{3}{c|}{1e-3 (relative), 1e-4 (absolute)} & \multicolumn{3}{c|}{1e-5 (relative), 1e-6 (absolute)} \\ \hline
\end{tabular}
\label{tab:hyper-param}
\end{table}

All hyperparameters are shown in Table \ref{tab:hyper-param}. Specifically, 
\begin{itemize}
    \item ``Learning rate'' refers to the learning rate for the Adam optimizer to train the dynamics model.
    \item ``GRU'' refers to the architecture of the GRU in all experiments, including the encoder in the Latent-RNN and Latent-ODE;
    \item For encoder-decoder models, we use a neural network with one layer and Tanh activation to convert the final hidden state of the encoder to the mean and log variance (for applying reparameterization trick to train VAE) of the initial latent state of the decoder. ``Encoder hidden to latent'' refers to the number of hidden units of this neural network.
    \item We use a neural network with one layer and Tanh activation for the interval timer $g_\kappa$. ``Interval timer $g_\kappa$'' refers to the number of hidden units of this neural network.
    \item We use a neural network with two layers and Tanh activation for the ODE network $f_\theta$. ``ODE network $f_\theta$'' refers to the number of hidden units of this neural network.
    \item ``ODE solver'' refers to the numerical ODE solver we use to solve the ODE (Equations \ref{eq:ode-rnn} and \ref{eq:latent-ode}). Note that we do not use the adjoint method \citep{chen2018neural} for ODE solvers due to a longer computation time. 
    \item ``ODE error tolerance'' refers to the error tolerance we use to solve the ODE numerically.
\end{itemize}

\subsection{Policy}
\paragraph{DQN hyperparameters.} 
The DQN for three simpler domains has two hidden layers of 256 and 512 units each with Relu activation. Parameters are trained using an Adam optimizer with a learning rate 5e-4 and a batch size of 128. We minimize the temporal difference error using the Huber loss, which is more robust to outliers when the estimated Q-values are noisy. We update the target network every 10 episodes (hard update). The action is one-hot encoded as the input for DQNs. To improve the performance of the DQN, we use a prioritized experience replay buffer \citep{schaul2015prioritized} with a prioritization exponent of 0.6 and an importance sampling exponent of 0.4, and its size is 1e5. To encourage exploration, we construct an $\epsilon$-greedy policy with an inverse sigmoid decay scheme from 1 to 0.05. Also, all final policies are softened with $\epsilon=0.05$ for evaluation. 

\paragraph{DDPG hyperparameters.} 
The DDPG networks for both the actor and critic have two hidden layers of 64 units each with Relu activation. Parameters are trained using an Adam optimizer with a learning rate of 1e-4 for the actor, a learning rate of 1e-3 for the critic and a batch size of 128. The target networks are updated with the rate 1e-3. The size of the replay buffer is 1e6. To encourage exploration, we collect 15000 samples with a random policy at the beginning of training, and add a Gaussian noise $\mathcal{N}(0, 0.1^2)$ to every selected action. 

\paragraph{Discount factors.} The discount factor is 0.99 for the windy gridworld and Mujoco tasks, is 0.995 for the HIV domain, and is 1 for the acrobot problem.

\begin{figure}[t]
    \centering
    \includegraphics[width=\textwidth]{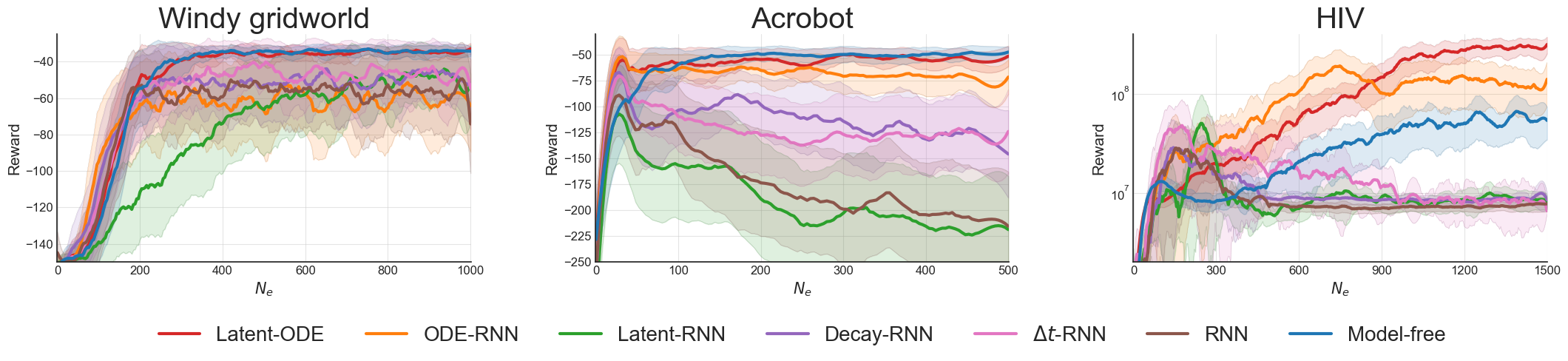}
    \caption{Learning curves with all baselines on three simpler domains. The $x$-axis is the number of episodes in Algorithm \ref{algo:world-model} (for the model-free baseline, the $x$-axis is the number of episodes in the actual environment). The shaded region represents a standard deviation of average evaluation over four runs (evaluation data is collected every 4 episodes). Curves are smoothed with a 20-point window for visual clarity.} 
    \label{fig:reward-world}
\end{figure}

\section{Additional figures and tables}\label{apx:res}
\subsection{Learning curves of learning world models}
\begin{wrapfigure}{r}{0.4\textwidth}
\centering
\vspace{-0.1cm}
\includegraphics[width=0.4\textwidth]{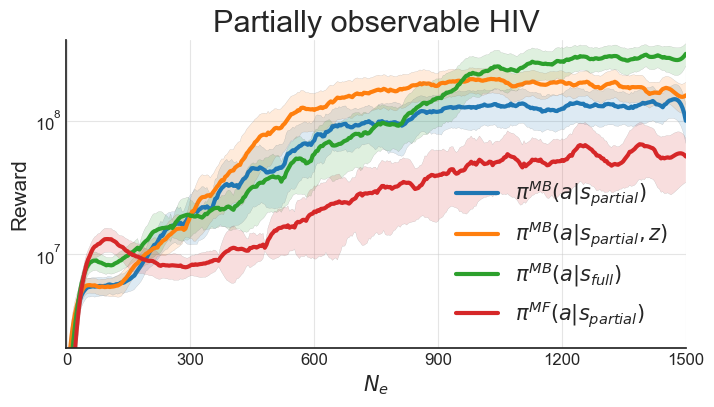}
\vspace{-0.65cm}
\caption{Learning curves of different policies on the partially observable HIV environment.}
\vspace{-0.3cm}
\label{fig:pomdp}
\end{wrapfigure}
Figure \ref{fig:reward-world} shows the learning process of all baselines on three simpler domains. Note that Figure \ref{fig:reward-world} does not necessarily demonstrate the sample efficiency because the model-free method uses the online real data, whereas the model-based approach (learning world models) uses the offline real data. However, we still observe that the Latent-ODE and ODE-RNN develop a high-performing policy much faster than the model-free baseline over 1500 episodes in the HIV environment. The model-free baseline converges after 3500 episodes. In addition, the acrobot problem clearly shows the model difference in terms of abilities of modeling the continuous-time dynamics. The Latent-ODE and ODE-RNN outperforms other models; The RNN and Latent-RNN, designed for discrete-time transitions, totally lose their ways in continuous-time dynamics, but their similar performance might suggest the limited impact of architecture (recurrent vs. encoder-decoder); The $\Delta t$-RNN and Decay-RNN also struggle on modeling continuous-time dynamics though they leverage time interval information in different ways. 

Moreover, Figure \ref{fig:pomdp} shows the learning process of different policies on the partially observable HIV environment. The latent policy $\pi^{MB}(a|s_\text{partial}, z)$ develops a better-performing policy more quickly than the vanilla model-based policy $\pi^{MB}(a|s_\text{partial})$ and the model-free policy $\pi^{MF}(a|s_\text{partial})$.

\subsection{Full results of changing time intervals}\label{apx:res-regular}
Table \ref{tab:transfer-reward-full} shows the cumulative rewards of policies learned on regular measurements using pretrained models from the original irregular time schedule on three simpler domains. We can see that the Latent-ODE surpasses other baseline models in most situations. 

\begin{table}[t]
    \scriptsize
    \centering
    \vspace{-0.3cm}
    \caption{The cumulative reward (mean $\pm$ std, over five runs) of all baselines for all time discretizations. (a) windy gridworld; (b) acrobot; (c) HIV.}
    \label{tab:transfer-reward-full}
    \begin{subtable}[h]{0.01\textwidth}
    \caption{}
    \end{subtable}%
    \hspace{0.5cm}
    \begin{subtable}[h]{0.95\textwidth}
        \setlength{\tabcolsep}{3.2pt}
        \begin{tabular}{ccccccccc}
        \toprule
        & & RNN & $\Delta t$-RNN & Decay-RNN & Latent-RNN & ODE-RNN & Latent-ODE & Oracle\\ \midrule
        & $\tau=1$ & \bftab -31.59 $\pm$ 1.68 & -43.21 $\pm$ 1.25 & \bftab -31.69 $\pm$ 1.36 & -32.22 $\pm$ 1.19 & -47.31 $\pm$ 1.46 & -34.24 $\pm$ 1.93 & \emph{-44.17 $\pm$ 0.80} \\
        & $\tau=2$ & \bftab -31.95 $\pm$ 2.10 & -39.37 $\pm$ 3.37 & -32.27 $\pm$ 2.39 & -32.75 $\pm$ 1.60 & -32.40 $\pm$ 1.43 & -33.82 $\pm$ 2.32 & \emph{-31.61 $\pm$ 1.60} \\
        & $\tau=3$ & -35.08 $\pm$ 4.91 & -34.49 $\pm$ 4.82 & -35.03 $\pm$ 4.87 & -43.74 $\pm$ 8.71 & -36.61 $\pm$ 7.04 & \bftab -32.74 $\pm$ 2.22 & \emph{-32.29 $\pm$ 1.60} \\  
        & $\tau=4$ & -38.55 $\pm$ 7.99 & -36.00 $\pm$ 5.41 & -40.58 $\pm$ 8.56 & -36.94 $\pm$ 5.73 & -64.10 $\pm$ 14.78 & \bftab -33.26 $\pm$ 1.86 & \emph{-32.81 $\pm$ 1.74} \\  
        & $\tau=5$ & -44.33 $\pm$ 9.77 & -42.97 $\pm$ 9.43 & -43.55 $\pm$ 9.50 & -36.87 $\pm$ 5.04 & -87.50 $\pm$ 18.47 & \bftab -33.84 $\pm$ 2.02 & \emph{-33.35 $\pm$ 2.03} \\  
        & $\tau=6$ & -43.14 $\pm$ 8.75 & -54.78 $\pm$ 13.35 & -53.76 $\pm$ 13.68 & -42.95 $\pm$ 9.08 & -99.36 $\pm$ 16.46 & \bftab -35.76 $\pm$ 2.72 & \emph{-34.17 $\pm$ 2.09} \\  
        & $\tau=7$ & -61.01 $\pm$ 10.03 & -64.55 $\pm$ 10.89 & -60.78 $\pm$ 10.03 & -52.32 $\pm$ 8.91 & -114.70 $\pm$ 11.65 & \bftab -49.31 $\pm$ 6.62 & \emph{-35.93 $\pm$ 1.95} \\  
        \bottomrule 
        \end{tabular}
    \end{subtable}%
    \vspace{0.3cm}
    \begin{subtable}[h]{0.01\textwidth}
    \caption{}
    \end{subtable}%
    \hspace{0.5cm}
    \begin{subtable}[h]{0.95\textwidth}
        \setlength{\tabcolsep}{1.5pt}
        \begin{tabular}{ccccccccc}
        \toprule
        & & RNN & $\Delta t$-RNN & Decay-RNN & Latent-RNN & ODE-RNN & Latent-ODE & Oracle\\ \midrule
        & $\tau=0.2$ & -407.46 $\pm$ 13.82 & -281.92 $\pm$ 9.99 & -285.07 $\pm$ 8.47 & -237.25 $\pm$ 10.29 & -190.82 $\pm$ 9.13 & \bftab -171.37 $\pm$ 10.07 & \emph{-78.75 $\pm$ 3.23}\\
        & $\tau=0.4$ & -268.36 $\pm$ 15.34 & -199.72 $\pm$ 12.39 & -268.33 $\pm$ 15.50 & -258.23 $\pm$ 9.60 & -128.93 $\pm$ 12.48 & \bftab -82.24 $\pm$ 8.01 & \emph{-48.76 $\pm$ 2.04} \\
        & $\tau=0.6$ & -150.10 $\pm$ 11.59 & -168.28 $\pm$ 10.33 & -154.53 $\pm$ 11.72 & -232.42 $\pm$ 9.49 & -106.56 $\pm$ 10.74 & \bftab -68.58 $\pm$ 8.62 & \emph{-47.30 $\pm$ 3.95} \\  
        & $\tau=0.8$ & -143.98 $\pm$ 8.64 & -145.16 $\pm$ 6.57 & -180.71 $\pm$ 8.48 & -188.45 $\pm$ 8.54 & -72.00 $\pm$ 9.84 & \bftab -63.81 $\pm$ 8.47 & \emph{-38.18 $\pm$ 3.15} \\  
        & $\tau=1$ & -147.25 $\pm$ 9.55 & -66.16 $\pm$ 11.29 & -138.52 $\pm$ 13.59 & -155.81 $\pm$ 8.30 & -51.78 $\pm$ 10.16 & \bftab -39.24 $\pm$ 7.24 & \emph{-23.14 $\pm$ 2.50} \\  
        \bottomrule 
        \end{tabular}
    \end{subtable}%
    \vspace{0.3cm}
    \begin{subtable}[h]{0.01\textwidth}
    \caption{}
    \end{subtable}%
    \hspace{0.5cm}
    \begin{subtable}[h]{0.95\textwidth}
        \setlength{\tabcolsep}{5pt}
        \begin{tabular}{ccccccccc}
        \toprule
        & & RNN & $\Delta t$-RNN & Decay-RNN & Latent-RNN & ODE-RNN & Latent-ODE & Oracle\\ \midrule
        & $\tau=1$ & 4.72 $\pm$ 0.55 & 2.25 $\pm$ 0.31 & 5.86 $\pm$ 0.70 & 7.02 $\pm$ 1.02 & 0.58 $\pm$ 0.07 & \bftab 7.73 $\pm$ 0.99 & \emph{176.75 $\pm$ 56.50}\\
        & $\tau=2$ & 7.28 $\pm$ 3.34 & 9.01 $\pm$ 3.29 & 8.76 $\pm$ 3.97 & 11.46 $\pm$ 5.07 & 2.80 $\pm$ 0.86 & \bftab 18.54 $\pm$ 7.10 & \emph{70.74 $\pm$ 23.61} \\
        & $\tau=3$ & 13.85 $\pm$ 4.84 & 20.10 $\pm$ 3.97 & 7.67 $\pm$ 3.47 & \bftab 26.60 $\pm$ 7.67 & 8.59 $\pm$ 2.72 & 4.95 $\pm$ 1.56 & \emph{40.32 $\pm$ 5.56} \\  
        & $\tau=4$ & 7.30 $\pm$ 2.58 & \bftab 22.76 $\pm$ 4.78 & 7.20 $\pm$ 3.26 & 21.52 $\pm$ 4.59 & 12.30 $\pm$ 7.93 & \bftab 22.14 $\pm$ 2.32 & \emph{35.58 $\pm$ 2.47} \\  
        & $\tau=5$ & 7.66 $\pm$ 1.79 & 17.21 $\pm$ 2.44 & 5.84 $\pm$ 1.62 & 16.95 $\pm$ 3.05 & 11.32 $\pm$ 1.09 & \bftab 21.60 $\pm$ 2.39 & \emph{33.55 $\pm$ 1.97} \\
        & $\tau=6$ & 5.36 $\pm$ 2.14 & 4.51 $\pm$ 1.41 & 3.52 $\pm$ 1.49 & 11.74 $\pm$ 3.20 & 7.82 $\pm$ 3.37 & \bftab 16.67 $\pm$ 2.79 & \emph{19.74 $\pm$ 0.94}\\
        & $\tau=7$ & 2.09 $\pm$ 0.77 & 7.34 $\pm$ 2.02 & 2.34 $\pm$ 0.97 & \bftab 9.83 $\pm$ 3.11 & 2.45 $\pm$ 0.20 & 8.21 $\pm$ 1.90 & \emph{12.33 $\pm$ 0.91} \\
        \bottomrule 
        \end{tabular}
    \end{subtable}
\end{table}

\end{appendices}

\end{document}